\documentclass[prl,twocolumn,superscriptaddress]{revtex4-1}

\usepackage{amssymb}
\usepackage{amsmath}
\usepackage{bbold}

\usepackage{braket}
\usepackage{xcolor}

\usepackage{graphicx}
\graphicspath{{../figura1/}{../figura2/}}

\begin{document}


\title{Beyond the storage capacity: data driven satisfiability transition}

\author{Pietro Rotondo}
\affiliation{I.N.F.N Milano}
\affiliation{Universit\`a degli Studi di Milano}

\author{Mauro Pastore}
\affiliation{Universit\`a degli Studi di Milano}
\affiliation{I.N.F.N Milano}

\author{Marco Gherardi}
\affiliation{Universit\`a degli Studi di Milano}
\affiliation{I.N.F.N Milano}

\begin{abstract}
Data structure has a dramatic impact on the properties of neural networks, yet its significance in the established theoretical frameworks is poorly understood.
Here we compute the Vapnik-Chervonenkis entropy of a kernel machine operating on data grouped into equally labelled subsets.
At variance with the unstructured scenario, entropy is non-monotonic in the size of the training set, and displays an additional critical point besides the storage capacity.
Remarkably, the same behavior occurs in margin classifiers even with randomly labelled data,
as is elucidated by identifying the synaptic volume encoding the transition.
These findings reveal aspects of expressivity lying
beyond the condensed description provided by the storage capacity,
and they indicate the path towards more realistic bounds for the generalization error
of neural networks.
\end{abstract}

\maketitle

\paragraph{Introduction---} 
The success of deep learning has transformed data science profoundly in the last decade,
within and outside physics \cite{LeCun2015, Goodfellow2016, He_2016_CVPR}.
In spite of the accomplishments in practical applications, we are currently facing
a lack of fundamental theoretical understanding in the field 
\cite{Mallat2016, BaldassiBorgs:2016}.
Outstanding open questions concern the surprising effectiveness
of stochastic gradient descent,
which is capable of finding good minima in complex energy landscapes,
and the identification of informative metrics
to predict the performances of
deep (many small layers) and shallow (few large layers) neural networks
\cite{Raghu2017,MeiE7665, ChaudhariSoatto:2018}.
Particularly troublesome is the apparent incompatibility,
within the accepted mathematical theories,
between the expressive power and the generalization abilities of neural networks:
ultimately, the reason why deep architectures with millions of parameters 
generalize well is mostly unknown
\cite{ZhangBengio2017,MartinMahoney2017,Chaudhari2019,Neyshabur2017,BoSaad2018}.

A natural frame for these issues is statistical learning theory \cite{vapnik2013nature},
which provides upper bounds to the probability of observing a large
generalisation error from a learning model with a given complexity.
These bounds are often distribution independent, i.e, they
are uniform in the generative model for the training data.
The downside of their universality
is their tendency to be too loose to be useful in practice.
New measures of complexity are being studied to fill this gap, and the urgency of
formulating data-dependent theories
is widely expressed in the computer science literature 
\cite{Bottou2015,Antos2003,Lugosi2001,Shawe-Taylor1998,Cohn1992}.

While mathematical bounds usually address worst-case generalization,
the main originality of the statistical physics approach is the analysis of the typical case;
the distribution of the training data is therefore always an explicit ingredient of the computations.
However, since the classic work of Gardner \cite{Gardner:1987}, data distribution has been regularly assumed to be
factorized between the inputs and their labels, thus leaving no room for 
their dependence,
which is in essence what we call ``data structure'' here.
This attitude is changing, and there is now a surge of interest towards the role of data in machine learning,
with the goal of quantifying the extent to which the specificities of a data set
affect the performance of data-science methods and learning algorithms
\cite{ChungLeeSompolinsky:2016,ChungCohen:2018,ChungLeeSompolinsky:2018, Cohen2020, goldt2019modelling, gerace2020generalisation, erba2020random, Erba2019,Borra:2019}.
The main objective of this Letter is to investigate the effect that data structure has on the model complexity
of simple architectures in machine learning.
Previous research in the physics literature addressed this question via the traditional concept
of storage capacity $\alpha_\mathrm{c}$, which measures the maximum load $\alpha$
(number of data points over number of parameters)
that a model can learn with probability 1 in the thermodynamic limit.
By viewing supervised learning as a constraint satisfaction problem,
capacity corresponds to the transition between a satisfiable (SAT)
and an unsatisfiable (UNSAT) phase,
above which perfect training accuracy is achievable with probability $0$.
Here we show that the compact description of learning provided by the capacity 
hides important detail about the model, related to its expressive power on 
structured data.
Our point of originality is the shift from the capacity to a quantity borrowed
from the foundations of statistical learning theory: the Vapnik-Chervonenkis (VC) entropy.
We show that the VC entropy is non-monotonic as a function of the load,
and vanishes asymptotically, at variance with the data-agnostic setting.
This also contrasts with the classic bounds in statistical learning theory,
which are mostly obtained by upper bounding the VC entropy 
with quantities that grow polynomially in the size of the training set
\cite{Bousquet2004,Vapnkik1999}.
The hallmark of this non-monotonic behavior is an additional phase transition above the storage capacity.
The new critical point signals the entrance into the UNSAT phase of another
satisfiability problem, related to data structure.

\paragraph{Cover's computation---}
The VC entropy measures the expressive power of a classifier via the number of distinct dichotomies
of the input data that the model can represent.
A dichotomy is a function taking values in $\{0,1\}$; 
equivalently, it is a classification of the input data in two groups.
In principle, the VC entropy could give rise to informative bounds on the generalization error
(the average number of errors on the test set),
but it is usually very difficult to compute explicitly, thus statistical learning theory
resorts to more accessible complexity measures.

\begin{figure}[t]
\includegraphics[scale=1.14]{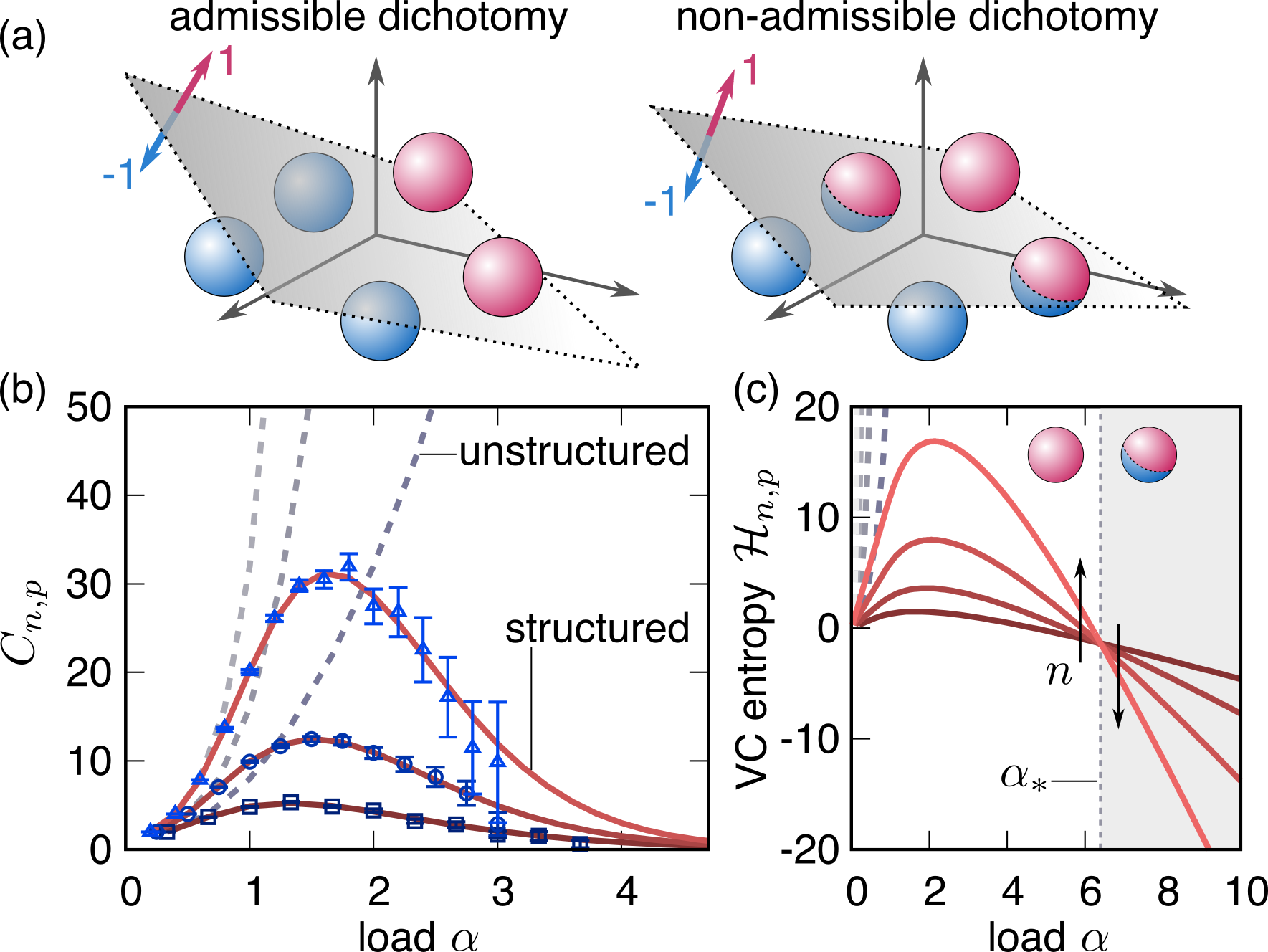}
\caption{
(a) 
Input data are structured as groups of points (here depicted by spheres)
sharing the same label (pink $=+1$, blue $=-1$).
(b-c)
The VC entropy $\mathcal{H}_{n,p}$ is the logarithm of the number $C_{n,p}$ of 
expressible dichotomies such that
no two points belonging to the same group are classified differently.
$C_{n,p}$ and $\mathcal{H}_{n,p}$ are
monotonic in the load for unstructured data and non-monotonic for structured data.
Solid lines are the theory for pairs of points ($k=2$); dashed lines
are Cover's result ($k=1$); from bottom to top, $n=3,4,5$ in (b) and $n=5,10,20,40$ in (c);
symbols are numerical estimates.
The VC entropies at different values of $n$ intersect roughly at
the same load $\alpha_*$, which separates two phases,
where admissible dichotomies are asymptotically present or absent.
}
\label{fig:1}
\end{figure}

Kernel architectures are a notable exception.
Their VC entropy was obtained analytically in a remarkable paper by Cover more than half a century ago \cite{cover1965}.
Cover calculated the number $C_{n,p}$ of dichotomies
as a function of the number $p$
of data points and the dimension $n$;
the VC entropy is $\mathcal H_{n,p}=\log C_{n,p}$.
In the thermodynamic limit, i.e., $n,p\to\infty$ with fixed load $\alpha=p/n$,
the fraction of dichotomies $C_{n,p}/2^p$ is discontinuous
at the storage capacity $\alpha_\mathrm{c}$
($\alpha_\mathrm{c}=2$ for the spherical Perceptron).
Remarkably, Cover's formula holds on very mild assumptions on the actual data points;
this suggests that statistical dependence between the inputs and their labels must be conceded
if one is to attain data-aware estimates.
Very recently the combinatorial technique devised by Cover was extended
to include this type of data structure \cite{Rotondo:2020:PRR},
allowing the computation of the number of ``admissible'' dichotomies, i.e., those that are compatible
with the data structure (see Fig.~\ref{fig:1}a). 

\paragraph{VC entropy in a simple model of data structure--- }

How to formulate a significant notion of data structure is a debated issue,
and different descriptions are useful in different contexts \cite{ChungLeeSompolinsky:2016, ChungLeeSompolinsky:2018, Mezard:2017,Mazzolini:2018:PRX}.
Here we use the definition of \cite{Rotondo:2020:PRR}.
Data points are grouped into $p$ subsets of $k$ points each,
where the labels are the same within each subset, 
and the geometric relations between points in a subset are fixed.
More precisely, the input set is 
$\Xi=\cup_{\mu=1}^p \Xi_\mu$, where each  $\Xi_\mu=\{\xi_a^{\mu}\}_{a=1,\ldots,k}$
is a set (``multiplet'') of $k$ points on the unit sphere 
$\xi_a^\mu\in S^{n-1}\subset \mathbb R^n$ 
such that their $k(k-1)/2$ overlaps are fixed:
$\xi_a^\mu \cdot \xi_b^\mu = \rho_{ab}$ for all $\mu=1,\ldots,p$.
The ensemble we consider is the flat probability measure on the 
$kp$ points $\xi_a^\mu$, conditioned to these constraints.
The admissible dichotomies $\phi$ of $\Xi$ are those for which
$\phi(\xi_a^\mu)=\phi(\xi_b^\mu)$ for all $a,b=1,\ldots,k$ and $\mu=1,\ldots,p$.
The usual unstructured ensemble is recovered either when $k=1$ (where no overlaps need to be specified),
or, for any $k$, when $\rho_{ab}=1$ for all $a,b$.
This model of data structure is closely related to the concept of ``perceptual manifolds''
inspired by neuroscience \cite{ChungLeeSompolinsky:2018,Seung:2000}, 
and was recognized in \cite{Borra:2019}
as a promising theoretical tool to address the problem of generalization.

The average number of admissible dichotomies $C_{n,p}$ of $p$ sets of $k$ points
(the logarithm of which is the VC entropy $\mathcal{H}_{n,p}$) 
satisfies the mean-field recurrence relation \cite{Rotondo:2020:PRR}
\begin{equation}
\label{eq:recurrence_k}
C_{n,p+1}=\sum_{l=0}^k \theta_l^k\, C_{n-l,p}.
\end{equation}
The boundary conditions depend mildly on the geometry, but they can be approximated by
$C_{n\geq 1,1}=2, C_{0,p}=0$.
Each coefficients $\theta_l^k$ in Eq.~(\ref{eq:recurrence_k}) depends on $k-1$ numbers 
$\{\psi_m\}_{m=2,\ldots,k}$, with $0\leq\psi_m\leq 1$,
having the following geometric-probabilistic interpretation.
Let $w\in S^{n-1}$ be a random vector with the uniform measure on the unit sphere.
Consider any multiplet $\Xi_\mu$, and a subset
$\Xi'\subseteq\Xi_\mu$ of $m\leq k$ points.
Then $\psi_m$ is the symmetrized probability that the scalar product $w\cdot\xi$
has the same sign for all $\xi\in\Xi'$, conditioned on it having the same sign
for all $\xi\in\Xi'\setminus\{\xi_\star\}$:
$\psi_m = 2 \left<\mathrm{Pr}\left[ (w\cdot \xi_\star) > 0 \;|\; (w\cdot \xi) > 0\; 
\forall \xi\in\Xi'\setminus\{\xi_\star\} \right]\right>_{\mathrm{sym}}$,
where the symmetrization $\left<\cdot\right>_\mathrm{sym}$ is performed by averaging over all subsets $\Xi'$
and over all choices of $\xi_\star\in\Xi'$.
These quantities can be expressed in terms of the overlaps $\rho_{ab}$,
e.g., $\psi_2(\rho)=2\pi^{-1}\arctan \sqrt{(1+\rho)/(1-\rho)}$.
%
%

Remarkable differences between structured and unstructured data
appear if one compares numerical solutions of Eq.~(\ref{eq:recurrence_k})
for $k=1$ (unstructured) and $k=2$ (structured) (Fig.~\ref{fig:1}).
The VC entropy $\mathcal{H}_{n,\alpha n}$, as a function of $\alpha$ at fixed $n$,
diverges with $\alpha$ in the unstructured case
(it does logarithmically, thus
the fraction of realizable dichotomies $C_{n,\alpha n}/2^{\alpha n}$ converges to $0$ for $\alpha\to\infty$).
On the contrary, $\mathcal{H}_{n,\alpha n}$ is
non-monotonic in the load for structured data,
and $C_{n,\alpha m}$ is itself asymptotically $0$.
Strikingly, curves corresponding to different values of $n$
cross each other roughly at the same load $\alpha_*$, similarly to
what $C_{n,\alpha n}/2^{\alpha n}$ does around the storage capacity $\alpha_\mathrm{c}$.
Hence, in the thermodynamic limit the VC entropy diverges to $+\infty$ for fixed $\alpha<\alpha_*$
and to $-\infty$ for $\alpha>\alpha_*$.
As will be elucidated by the following computations, this transition
is driven by a tradeoff between
an entropic term, related to the combinatorial growth of the number of dichotomies with the load,
and an energetic term, due to the constraints that define data structure.

\paragraph{Transition point via combinatorial analysis---}

The transition point in the thermodynamic limit is accessible by a perturbative analysis.
In some cases it is possible to solve Eq.~(\ref{eq:recurrence_k}) explicitly,
but we construct here an indirect method, based on analytic combinatorics
\cite{flajolet2009analytic}.
This method has the crucial advantage of being applicable despite
the fact that
(i) $C_{n,p}$ is not known in closed form for generic $k$, and (ii) the recurrence equation
itself has implicitly-defined coefficients
(see \cite{companion2} for details on the computations, and for applications to simpler cases
where the above restrictions do not apply).

Let $g_n(z)$ be the ordinary generating function of $C_{n,p}$ with respect to the variable $p$:
$g_n(z)=\sum_{p=1}^\infty C_{n,p} z^p$.
At fixed $n$, $g_n(z)$ encodes the large-$p$ asymptotics of $C_{n,p}$ via its singular behavior.
In particular, if $g_n(z)$ is a rational function the dominant pole of which
is of order $r$ and lies at $z=z_0$, with finite part 
$R \equiv \lim_{z\to z_0} (z_0-z)^r g_n(z)$, then, for large $p$,
$C_{n,p} \sim R\, z_0^{-p-r} B(p+r-1,r-1)$,
where $B(a,b)$ is the binomial coefficient ${a}\choose{b}$.

Multiplying Eq.~(\ref{eq:recurrence_k}) by $z^p$ and summing over $p$
(taking care of the boundary conditions)
gives a recurrence relation for $g_n(z)$:
\begin{equation}
\label{eq:recurrence_gn}
g_n(z)=\frac{z}{1-z \theta^k_0} \left[ 2+ \sum_{l=1}^k \theta^k_l g_{n-l}(z)\right],
\end{equation}
with $g_{n\leq 0}(z)=0$.
Iteration of Eq.~(\ref{eq:recurrence_gn}) $n$ times, starting from the non-singular initial condition at $n=0$,
yields a singular $g_n(z)$, whose pole, generated by the pole in the right-hand side of the recurrence relation,
lies at $z_0=1/\theta^k_0$, has order $r=n$, and finite part 
$R=2(\theta^k_1)^{n-1}(\theta^k_0)^{-2n}$.
Finally, the asymptotic form of the VC entropy is
$\mathcal{H}_{n,\alpha n} \sim \log C(\alpha;n)$, with
\begin{equation}
\label{eq:C_alpha}
C(\alpha;n) = 2 \frac{\Gamma(\alpha n+n)}{\Gamma(n)\Gamma(\alpha n+1)} 
(\theta^k_1)^{n-1} (\theta^k_0)^{(\alpha-1)n}.
\end{equation}
Conveniently, $C(\alpha;n)$ depends only on the first two $\theta^k_l$'s 
(see \cite{companion2} for their expressions as functions of the probabilities $\psi_m$).
The transition is at the point $\alpha=\alpha_*$ where the VC entropy
is asymptotically constant in $n$, i.e.,
$\partial_n \mathcal{H}_{n,\alpha_* n} \to 0$.
From Eq.~(\ref{eq:C_alpha}) one obtains
\begin{equation}
\label{eq:alpha_star}
S(\alpha_*)
+(\alpha_*-1)\log\theta^k_0 +\log\theta^k_1 = 0,
\end{equation}
with $S(\alpha)\equiv(\alpha+1)\log(\alpha+1)-\alpha\log\alpha$.
Equation (\ref{eq:alpha_star}) expresses the tradeoff between a positive entropic term $S(\alpha)$,
the same as for unstructured data, and a structure-dependent energetic term.
It has two solutions: $\alpha_*$ is the larger.

\begin{figure}[tb]
\includegraphics[scale=1.14]{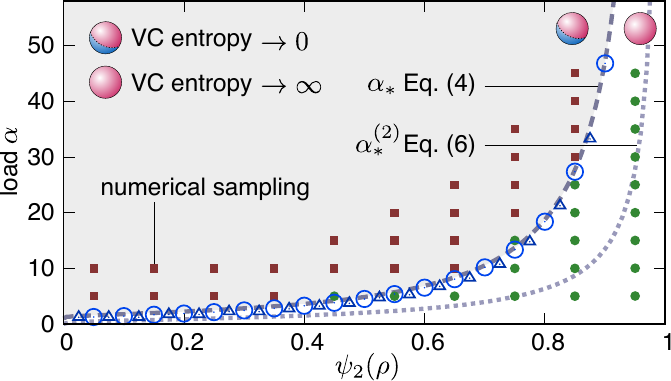}
\caption{
Phase diagram of the VC entropy for $k=2$.
The dashed line is the theoretical prediction for $\alpha_*$ obtained by combinatorial methods;
the dotted line is the transition line of the synaptic volume \eqref{Gardner}
in the annealed approximation.
Empty symbols are numerical results, obtained by finding the intersection
between two curves $C_{n_1,\alpha n_1}$ and $C_{n_2,\alpha n_2}$ 
with $n_1=40, n_2=20$ (circles)
and $n_1=6,n_2=3$ (triangles);
each filled symbol is obtained by sampling $10^5$ random inputs
with $n=3$
(red squares = no admissible dichotomy,
green circles = at least one admissible dichotomy).
}
\label{fig:2}
\end{figure}

Consider the case $k=2$, where input data are pairs of points with fixed pairwise overlap $\rho$.
Then $\theta^k_0=\psi_2(\rho)$, $\theta^k_1=1$, and $\alpha_*$
is an increasing function of $\rho$.
Coherently, $\alpha_*$ diverges when $\rho\to 1$, thus recovering
the unstructured case $k=1$, where no transition is present.
Figure \ref{fig:2} shows that (i) the value of $\alpha_*$
satisfying Eq.~(\ref{eq:alpha_star}) matches that
obtained by numerical integration of the recursion Eq.~(\ref{eq:recurrence_k}),
and (ii) the transition can be probed by sampling small random linear classifiers
(see the caption).
The phenomenology is the same for all $k$ \cite{companion2}.
Similarly to the critical behavior at the SAT-UNSAT transition of random constraint satisfaction problems
\cite{Kirkpatrick1994, Leone2001},
the number of dichotomies, as a function of the reduced control parameter $\hat\alpha = (\alpha-\alpha_*)/\alpha_*$,
obeys a finite-size scaling form
$C(\alpha;n) = n^{-\beta/\nu} F\left( \hat\alpha\, n^{1/\nu}  \right)$,
with critical exponents $\beta = 1/2$ and $\nu = 1$,
where $F$ is a regular function (see \cite{companion2} for the explicit formula).
At $\alpha=\alpha_*$, $C(\alpha;n)$ vanishes as a power law in the dimensionality $n$;
the exponent $\nu$ controls the scaling of the width of the critical region
(by contrast, $\nu=2$ at the storage capacity).
%

\paragraph{Identification of the relevant synaptic volume--- }

The phase transition at $\alpha_*$ can be interpreted as the SAT-UNSAT transition 
of the following constraint satisfaction problem:
given a realization of the ``disorder'' $\Xi$, find a vector $W$ identifying a linearly-realizable
dichotomy of $\Xi$ that is admissible.
This characterization indicates that the following synaptic volume should
pinpoint the transition:
\begin{equation}
\label{Gardner}
V\left(\Xi\right) =\int D^p \sigma \int D^n W \prod_{\mu,a = 1}^{p,k} \theta \left[\sigma^\mu \sum_{i=1}^n W_i (\xi_a^\mu)_i\right],
\end{equation}
where $\theta[\cdot]$ is the Heaviside theta,
$(\xi_a^\mu)_i$ denotes the $i$-th component of the $a$-th element of the $\mu$-th multiplet,
$D^n W$ is a shorthand for a Gaussian or spherical measure over the weights,
and $D^p \sigma =  \prod_\mu [ \delta(\sigma^\mu-1) + \delta(\sigma^\mu+1) ]\mathrm{d}\sigma^\mu$.
Besides the data structure, encoded in the multiplets $\Xi_\mu$,
the synaptic volume~(\ref{Gardner}) differs from the ordinary Gardner volume by the integration over the labels $\sigma$.
Intuitively, as long as $V(\Xi)$ grows exponentially with $n$ at fixed load $\alpha$,
at least one classification compatible with the input-label constraints can be expressed by the model.
Thus, the scaling of $V(\Xi)$ is a proxy of the nonmonotonic behavior
of the VC entropy for a given data structure.

We restrict the analysis to data structured as pairs of points ($k=2$),
and we compute $V(\Xi)$ in the simplest approximation scheme, averaging
at the annealed level over the inputs.
(See \cite{companion2} for the replica theory.)
For $\rho=1$ we recover the unstructured case:
$\left<V(\Xi)\right>$ diverges for any load $\alpha$,
in agreement with Cover's theory
(a polynomial number of classifications can be realised by a kernel architecture).
The situation changes for $\rho<1$.
In this regime data structure becomes relevant, and there appears a critical load 
$\alpha_*^{(2)} (\rho)$
for which the synaptic volume 
shrinks exponentially fast in $n$. Above this threshold, which is given by
\begin{equation}
\alpha_*^{(2)} (\rho)= - \frac{\log (2\pi) +1}{2 \log \left(1/2 +\pi^{-1} \arcsin \rho\right)},
\end{equation} 
none of the classifications compatible with the data structure can be realised by the kernel architecture.
The threshold computed in the annealed approximation provides a lower bound to 
the $\alpha_*$ evaluated by the combinatorial approach (see Fig. \ref{fig:2}).

\paragraph{Margin-driven transition with unstructured data---}

Margin classifiers are prominent in statistical learning theory,
as their generalization error can be kept under control via the margin,
and they lie at the core of the powerful idea of support vector machines
\cite{Cortes1995, 788640}.
A significant observation linking classification with margin
and classification of structured data was done in \cite{ChungLeeSompolinsky:2018}:
linear classification with margin $\kappa$ is equivalent 
to learning a set of spherical manifolds with radius equal to the margin.
The equivalence, valid for a kernel machine with kernel $\varphi$, holds in the following sense:
the set of $d$-dimensional weights
$W$ in feature space realizes the mapping with margin $\kappa$ if and only if $\sigma^\mu=\mathrm{sign}(W\cdot \zeta^\mu)$ for all $\mu$ and all $\zeta^\mu$ such that $\left|\zeta^\mu-\varphi(\xi^\mu)\right|^2<\kappa^2$.
Intuitively, the constraints of the satisfiability problem are shifted from
the data to the function class.
(If the margin is negative the problem is no more convex,
and bears connections to jamming phenomena \cite{FranzParisi2016}.)

This observation suggests that the VC entropy of a margin classifier with randomly labelled
(i.e., unstructured) data
should present the same phenomenology described above for data structured in multiplets.
To our knowledge, there is no combinatorial technique to compute the entropy in this case,
thus we use an integrated synaptic volume analogous to
Eq.~\eqref{Gardner} as a probe into the phase transition.
Again, in the annealed approximation,
the volume shrinks exponentially fast above a threshold load, given by
\begin{equation}
\alpha_*^\mathrm{M} (\kappa) = - \frac{\log (2 \pi) + 1}{2 \log\textrm{Erfc}(\kappa)}\,.
\end{equation}
As in the case of zero-margin classification of multiplets,
$\alpha_*^\mathrm{M}(\kappa)\to \infty$ when the constraints are relaxed ($\kappa\to 0$ in this case),
and $\alpha_*^\mathrm{M}(\kappa)\to 0$ when the constraints become unsatisfiable ($\kappa\to\infty$).

\paragraph{Discussion---} 
Finding compact scalar metrics 
descriptive of the complexity and the flexibility of a hypothesis space 
is a shared effort of statistical physics and statistical learning theory.
Unsophisticated quantities such as the number of degrees of freedom are merely superficial
indicators of the expressive power of a given model, and they fail at the task of characterizing
the model's generalization properties, especially in applications to non-synthetic data sets.
This is partly true even for more refined quantities such as the VC dimension and its distribution-dependent counterparts.
The importance of including data specificities in the existing frameworks is recognized
in both physics and computer science.
In particular, it is well appreciated that restricting the hypothesis class by imposing a margin
is beneficial to generalization.
A large body of work in modern SLT is devoted to prove data-dependent bounds on the generalization error.
However, these results are obtained by bounding the VC entropy with monotonically increasing functions
of the sample size $p$.
Our results suggest that, in principle, these results could be improved substantially
already by including rather unrestrictive priors on the data distribution.

Here, in the spirit of statistical physics,
we have focused on simple architectures and a simple implementation of data structure.
This approach enabled us to obtain tractable analytical expressions that serve,
in a wider context, as a proof of principle, and promote two main points:
(i) 
The concept of storage capacity in the statistical physics of machine learning
should be complemented by other, preferably data-oriented, ``order parameters'' of model complexity.
(ii)
Data structure, in the form of dependence or constraints between inputs and labels,
should be investigated in the framework of statistical learning theory,
acknowledging the possibility of an asymptotically decreasing VC entropy.
In this Letter we reported on the discovery of a data-driven phase transition,
which appears to be a good candidate for the pursuit of point (i).
Point (ii) is explored in more depth in \cite{companion2}.
How to address these issues for deep neural networks,
or even in more generality in the context of machine learning,
is compelling matter for future work.

\bibliographystyle{apsrev4-1}
\bibliography{biblio}

\end{document}